\documentclass[runningheads]{llncs}
\usepackage[T1]{fontenc}
\usepackage{graphicx}
\usepackage{amsmath,amssymb,amsfonts}
\usepackage{algorithmic}
\usepackage{graphicx}
\usepackage{textcomp}
\usepackage{xcolor}
\usepackage{subfig}
\usepackage{gensymb}
\usepackage[numbers]{natbib}
\usepackage{caption}
\usepackage{svg}
\usepackage{mwe}
\usepackage{amsfonts}
\usepackage{amssymb}
\usepackage{verbatim}
\usepackage{amsbsy}
\usepackage{siunitx}
\usepackage{tabularx, booktabs}
\def\BibTeX{{\rm B\kern-.05em{\sc i\kern-.025em b}\kern-.08em
    T\kern-.1667em\lower.7ex\hbox{E}\kern-.125emX}}

\usepackage{soul}
\usepackage{tikz}
\usetikzlibrary{calc}
\begin{document}
\title{UruBots UAV - Air Emergency Service Indoor Team Description
Paper for FIRA 2024}
%
%
\author{Hiago Sodre\inst{1}\and Sebastian Barcelona\inst{1} \and Anthony Scirgalea\inst{1} \and Brandon Macedo\inst{1} \and Gabriel Sampson\inst{1} \and Pablo Moraes\inst{1} \and William Moraes\inst{1} \and Victoria Saravia\inst{1} \and Juan Deniz\inst{1} \and Bruna Guterres\inst{1} \and Andre Kelbouscas\inst{1} \and Ricardo Grando\inst{1} }
\authorrunning{UruBots UAV et al.}
\titlerunning{UruBots UAV}
%
\institute{Technological University of Uruguay - UTEC}

\maketitle              
\begin{abstract}
This document addresses the description of the corresponding "Urubots" Team for the 2024 Fira Air League, "Air Emergency Service (Indoor)." We introduce our team and an autonomous Unmanned Aerial Vehicle (UAV) that relies on computer vision for its flight control. This UAV has the capability to perform a wide variety of navigation tasks in indoor environments, without requiring the intervention of an external operator or any form of external processing, resulting in a significant decrease in workload and manual dependence. Additionally, our software has been designed to be compatible with the vehicle's structure and for its application to the competition circuit. In this paper, we detail additional aspects about the mechanical structure, software, and application to the FIRA competition.

\end{abstract}
\section{Introduction}

In recent years, we have witnessed the growing use of Unmanned Aerial Vehicles (UAVs) in various fields such as industry, research, and military operations. This advancement has been made possible by the development of technologies such as 3D printing, detection, integrated processing, battery efficiency improvement, and miniaturization, which have enabled the creation of smaller and more efficient UAVs.

While most commercial UAVs are designed to operate in outdoor environments, using software systems that heavily rely on GPS-provided information to maintain their current position or navigate to new destinations, the availability and reliability of GPS indoors are limited. This has led to a shortage of UAVs capable of navigating indoors.

The proposed vehicle in this work is also designed to perform interior operations during a visual inspection, relying on visual information to locate and navigate autonomously. However, we focus o a vehicle with embedded processing, avoiding the use of offline processing. With this vehcile we aim to perform the tasks requested in the FIRA Air Emergency Service (Indoor) competition. Our UAV team has already competed in UAV categories, winning the second and third place in the Drone Hackathon of Uruguay 2022, a national competition organized by state companies.

\section{Construction}

\subsection{Hardware}

Starting with the drone's body, on the Figure 1 it is possible to view an image of the drone being built in its final stage ready for flight. This drone is composed of several components that make it autonomous, these components are represented in Figure ~\ref{fig:scenarios}.

\begin{figure}[httb]
     \centering
            \includegraphics[scale=0.2]{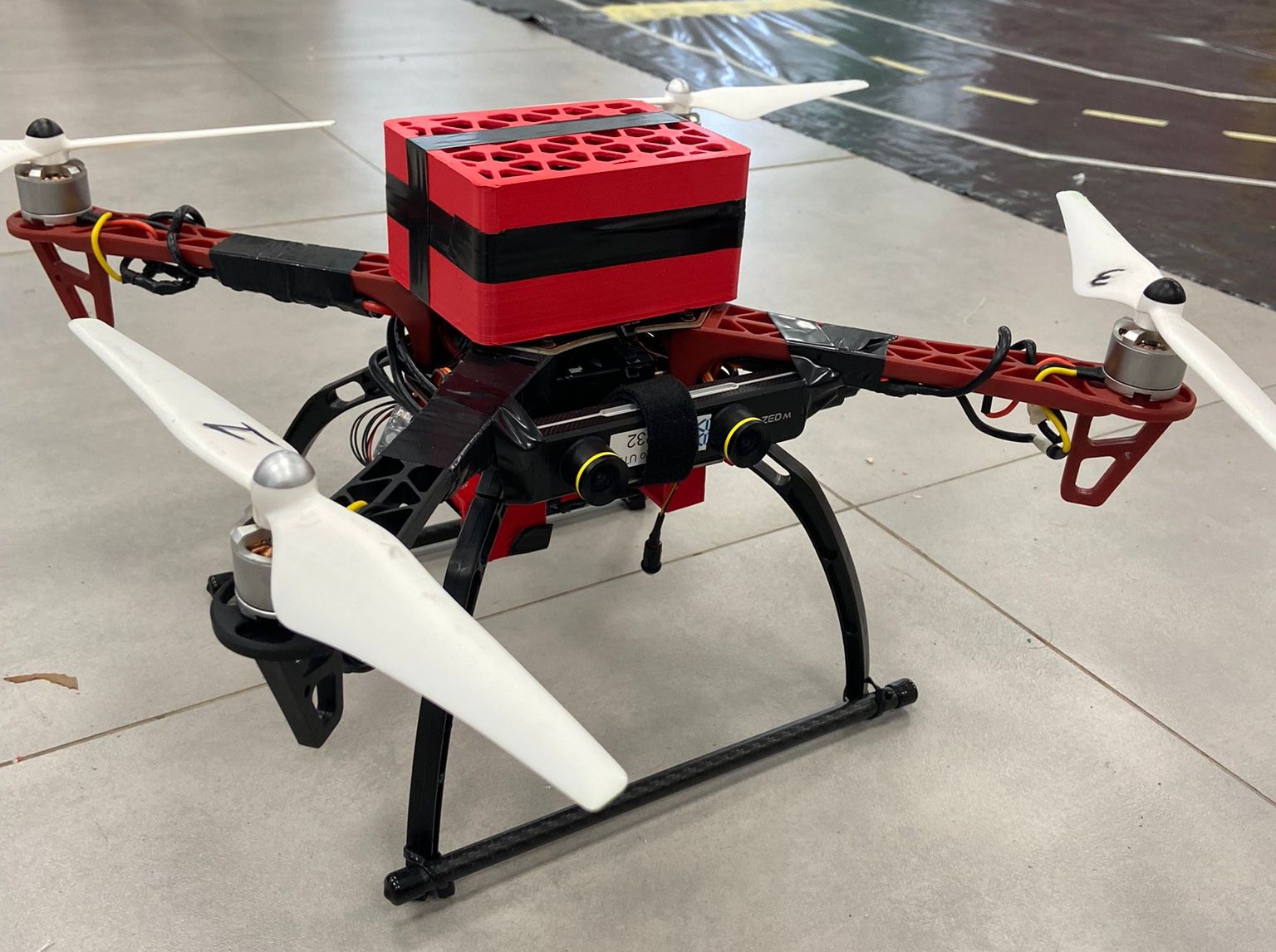}\label{fig:UAV}]
     \caption{UAV construction image.} 
\end{figure}

\begin{figure}[h]
  \subfloat[Frame F450.\label{fig:frame}]{
	\begin{minipage}[c][\width]{0.25\textwidth}
	   \centering
	   \includegraphics[width=\textwidth]{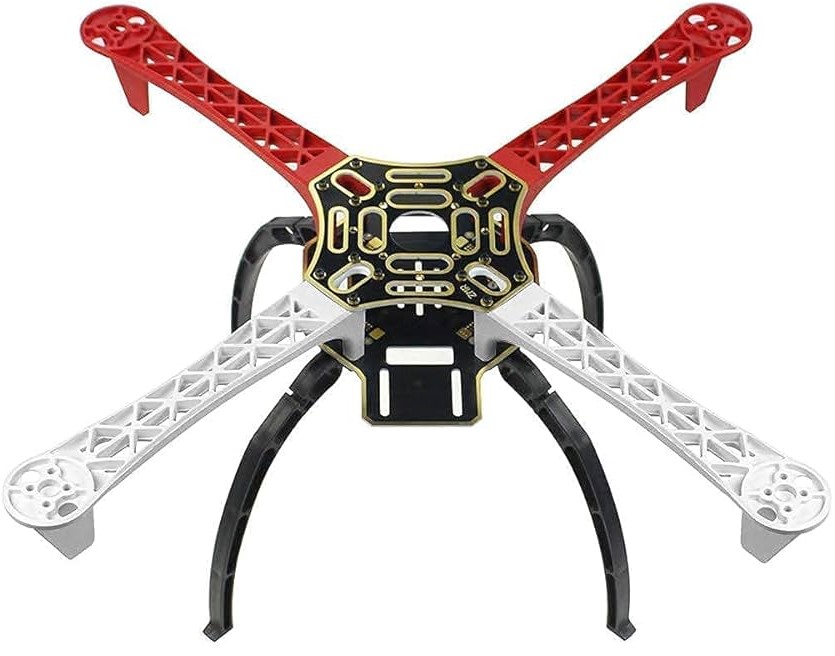}
	\end{minipage}} 
  \hfill 	
  \subfloat[Jetson Nano.\label{fig:jetson}]{
	\begin{minipage}[c][\width]{0.2\textwidth}
	   \centering
	   \includegraphics[width=\textwidth]{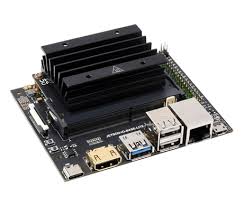}
	\end{minipage}}
  \hfill 	
  \subfloat[Battery 3s 2200mah.\label{fig:battery}]{
	\begin{minipage}[c][\width]{0.2\textwidth}
	   \centering
	   \includegraphics[width=\textwidth]{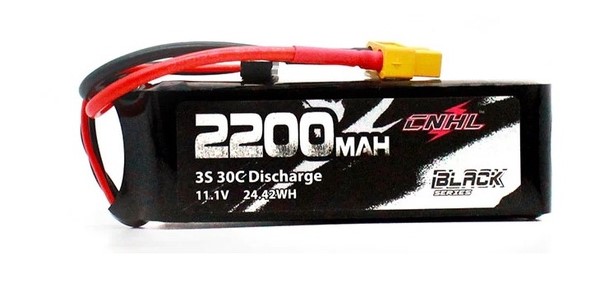}
	\end{minipage}}
  \hfill 	
  \subfloat[XL4016 Step Down Voltage Regulator Module.\label{fig:stepdown}]{
	\begin{minipage}[c][\width]{0.2\textwidth}
	   \centering
	   \includegraphics[width=\textwidth]{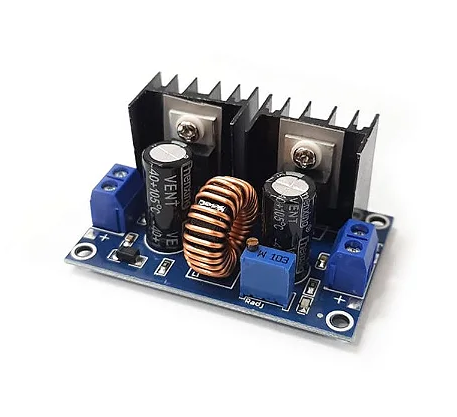}
	\end{minipage}}
  \hfill 	
  \subfloat[Mini ZED Camara.\label{fig:minized}]{
	\begin{minipage}[c][\width]{0.2\textwidth}
	   \centering
	   \includegraphics[width=\textwidth]{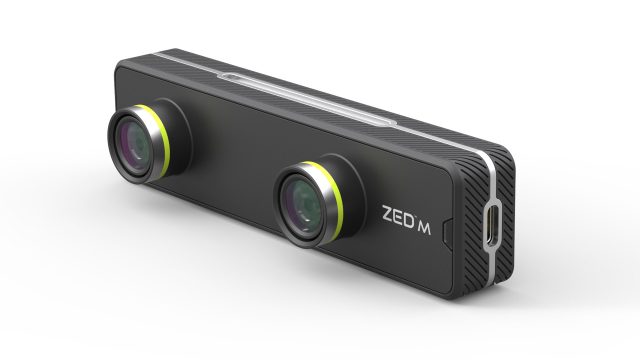}
	\end{minipage}}
  \hfill 	
  \subfloat[Logitech C270 camera.\label{fig:logitech}]{
	\begin{minipage}[c][\width]{0.2\textwidth}
	   \centering
	   \includegraphics[width=\textwidth]{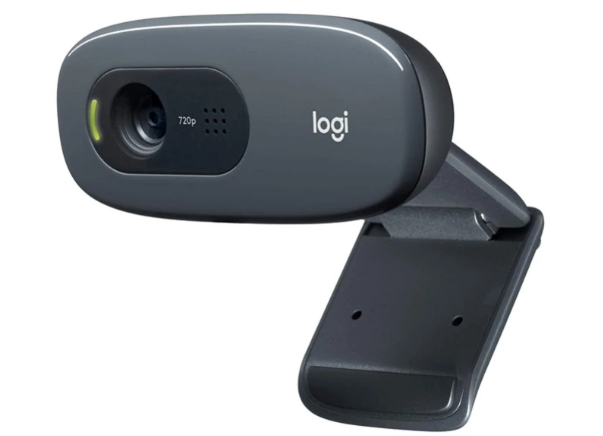}
	\end{minipage}}
  \hfill 	
  \subfloat[Pixhawk Board.\label{fig:pixhawk}]{
	\begin{minipage}[c][\width]{0.2\textwidth}
	   \centering
	   \includegraphics[width=\textwidth]{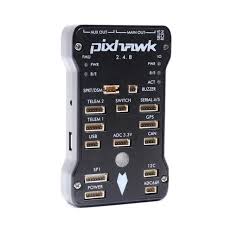}
	\end{minipage}}
  \hfill 	
  \subfloat[ESC PWM controller.\label{fig:esc}]{
	\begin{minipage}[c][\width]{0.2\textwidth}
	   \centering
	   \includegraphics[width=\textwidth]{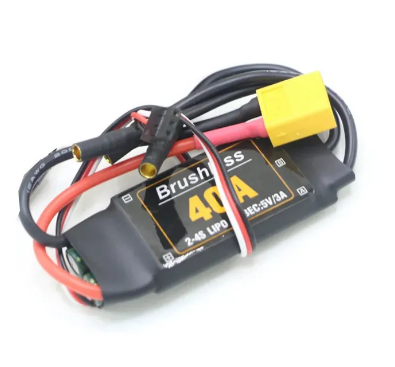}
	\end{minipage}}
  \hfill 	
  \subfloat[Motors.\label{fig:motor}]{
	\begin{minipage}[c][\width]{0.2\textwidth}
	   \centering
	   \includegraphics[width=\textwidth]{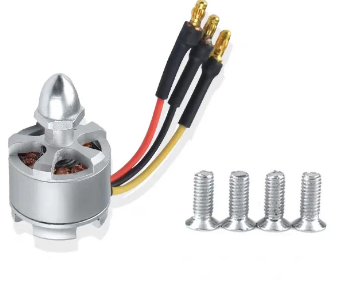}
	\end{minipage}}
  \hfill 	
  \subfloat[Propellers.\label{fig:helices}]{
	\begin{minipage}[c][\width]{0.2\textwidth}
	   \centering
	   \includegraphics[width=\textwidth]{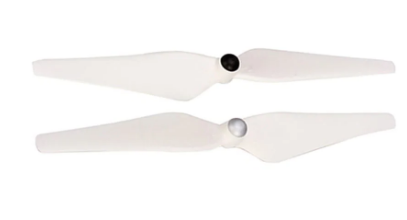}
	\end{minipage}}
  \hfill 	
  \subfloat[Wifi USB Adapter.\label{fig:tplink}]{
	\begin{minipage}[c][\width]{0.2\textwidth}
	   \centering
	   \includegraphics[width=\textwidth]{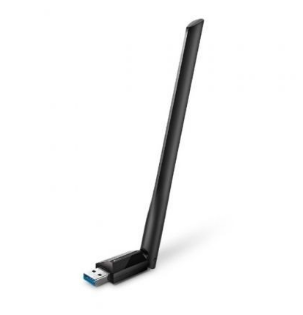}
	\end{minipage}}
\caption{Components of our UAV.}
\label{fig:scenarios}
\end{figure}

The first and main component of this autonomous system is the Jetson Nano processing board. It is a powerful and compact computing platform designed for machine learning projects. This board was chosen as it supports a high image rate processing required by the software due to its CUDA support and has a very low power consumption of around 10W. Jetson Nano is installed in the upper central part of the vehicle taking into account the center of mass of the vehicle, to comunicate with the computer we used a Tplink USB adaptor to give connection, his image can see in Figure ~\ref{fig:tplink}. A standard image of the Jetson Nano can be seen in Figure ~\ref{fig:jetson}. 

To power the system it was necessary to use Lipo type batteries (lithium polymer) which can be seen in Figure ~\ref{fig:battery}, with a nominal voltage of 11.1V (each cell has 3.7V) and three cells connected in series (3S), These batteries provide enough power for motors and electronic components. The 2200mAh capacity means it can deliver a current of 2200 milliamps for one hour.

As our Jetson Nano board does not work at the same operating voltage as the battery, it was necessary to use an XL4016 Step Down Voltage Regulator module, which is a compact and efficient device that converts a wide range of input voltages (4V to 40V) to a adjustable output voltage (1.25V to 36V). It can supply up to 8A of current, suitable for the maximum performance of our board. This board was installed alongside the Jetson Nano on top of the UAV, inside a case designed and printed in 3D to protect the components.

The chosen vision system was the Minized developed by the company StereoLabs. It is a compact, high-resolution stereoscopic camera designed for real-time computer vision applications. It can be seen in Figure ~\ref{fig:minized}. It was also necessary to use an additional camera to scan the Qr codes on the floor in a FIRA circuit task. To do this, we used a Logitech C270 in Figure ~\ref{fig:logitech}, in the lower part of the vehicle pointed to the ground.

Finally, to move the drone, we used four 2200kv brushless motors of the type shown in the ~\ref{fig:motor} to sopport the mass and to can fly this vehicle, and together to control the speed are four 40A escs like those in the ~\ref{fig:esc}. With this motors are used four propellers how like the \ref{fig:helices} to work with flight production.

\subsection{Software} 

The software system of our UAV is based on ROS and on the mavlink protocol. We created a mission planner ROS based on python that takes the information from the Pixhawk board to perform a sequence of tasks, such as navigating to a target position and to detect a qrcode.

\begin{figure}[httb]
     \centering
            \includegraphics[scale=0.4]{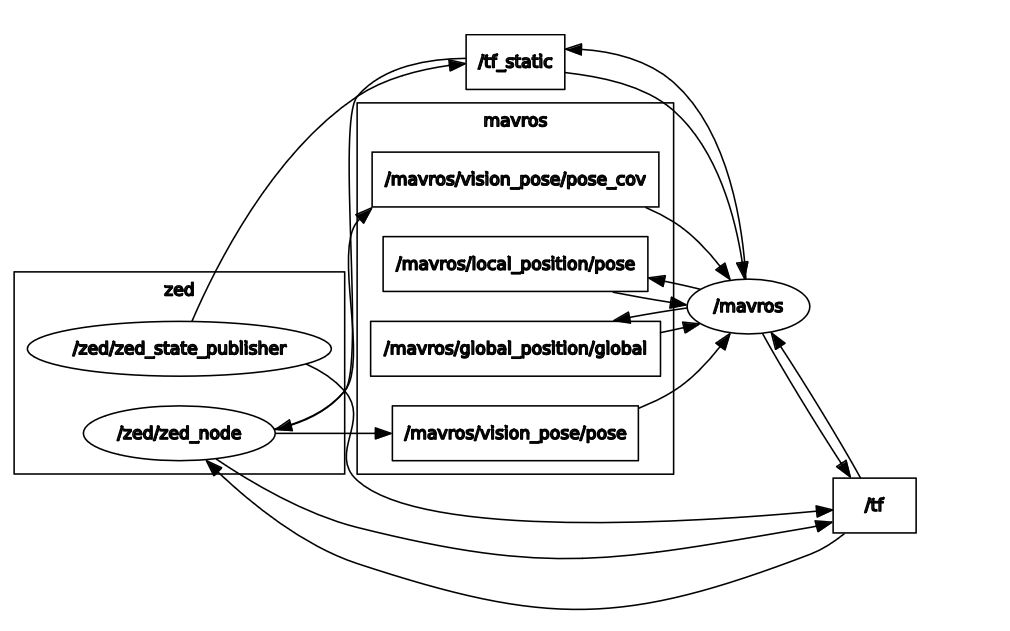}\label{fig:system_architecture}]
     \caption{UAV construction image.} 
\end{figure}

The ROS system also uses the mini ZED wrapper, which provides the camera's cartesian position. This info in forwarded to the Pixhawk board, which had its firmware changed so it can support a visual pose. Overall, the graph of our system can be seen in Figure \ref{fig:system_architecture}.

A package was developed for interface with Mavros to send commands to the vehicle, allowing it to have intelligence and work autonomously. A mission planning package was developed to send commands and obtain the vehicle status. Complementary packages to carry out environmental inspection were also developed to run on the Jetson Nano board along with the mission planner package. (Grando et al, 2020).

The detection system between two cameras was also a difficult task to implement due to the complexity of one camera always working to detect the Qrcode and giving command to the drone through the other camera if necessary.

\subsection{FIRA Challenge}

There are numerous situations in which the use of autonomous UAVs would be beneficial, such as inspecting damaged buildings, wind turbines, dense urban environments, or offshore platforms where GPS signal may be restricted. Currently, many UAVs in these scenarios rely on manual operation by a human operator, who primarily relies on video transmission to control the vehicle. However, this human dependency can result in difficulties in carrying out certain tasks, as well as potential communication failures that can lead to data loss, inaccurate navigation, or accidents.

The FIRA AIR competition emerges as a response to these challenges, incentivizing teams to develop vision-based autonomous UAVs with integrated intelligence, capable of performing stable flights, navigating, and locating themselves in indoor environments. Additionally, these UAVs are expected to advance by following routes based on reading QR codes, identifying QR codes on buildings, and autonomously navigating obstacles.

In this context, we have developed our vehicle, which does not require the intervention of an external operator or off-board processing, meeting the requirements set by the FIRA AIR competition. In figure 4a-b it is possible to visualize FIRA's idea of a competency scenario compared to the scenario replicated by the team in their work laboratory.

\begin{figure}[h]
  \subfloat[Competition scenario.\label{fig:comp_scenario}]{
	\begin{minipage}[c][0.55\width]{0.5\textwidth}
	   \centering
	   \includegraphics[width=\textwidth]{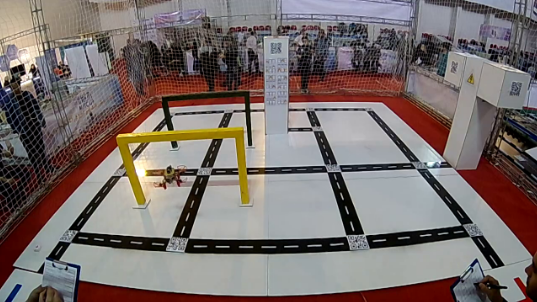}
	\end{minipage}} 
  \hfill 	
  \subfloat[Qualification scenario.\label{fig:qual_scenario}]{
	\begin{minipage}[c][0.55\width]{0.5\textwidth}
	   \centering
	   \includegraphics[width=\textwidth]{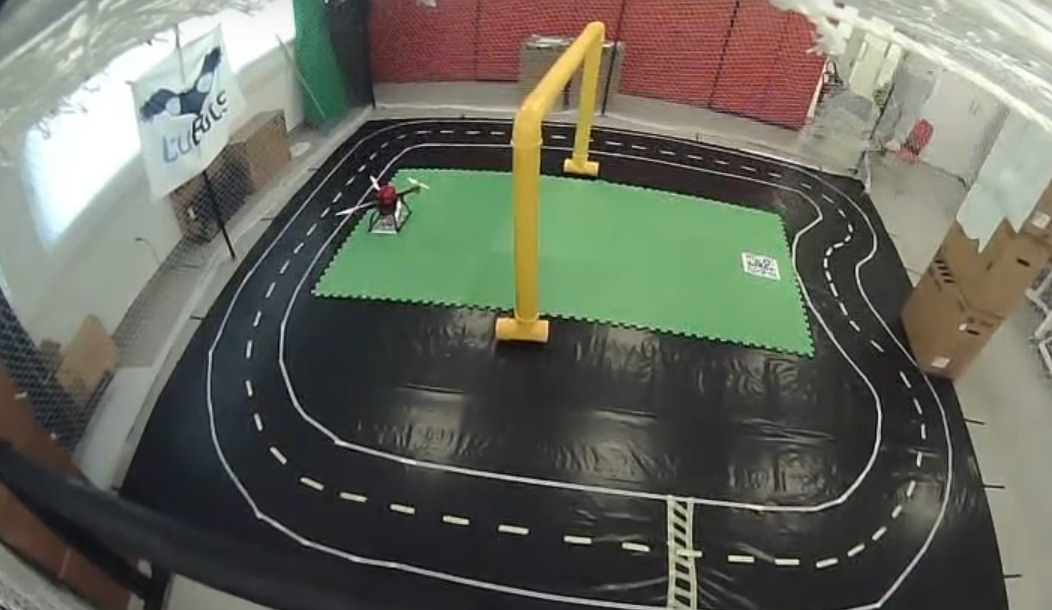}
	\end{minipage}}
\caption{Competition scenario and our qualification scenario.}
\label{fig:scenario}
\end{figure}

This scenario allows you to carry out different tests such as a safe flight within an ideal protection network, also with obstacles for complex flights and Qr codes for detection and free flight tests. The goal of the FIRA Air - Emergency Service competition is encouraging research teams to solve existing challenges in developing a smart and robust drone for both commercial and industrial applications. While drones are widely used for aerial imaging, there are still lots of challenges when it comes to an autonomous, reliable, and secure solution. Overall most of these difficulties are related to localization, exploration, and intelligent navigation in dynamic environments. (Farzad et al, 2023).




\end{document}